\documentclass{article}

\usepackage{natbib}
\usepackage{balance}

\usepackage{algorithm}
\usepackage{algorithmic}
\usepackage{amsmath}

\usepackage{booktabs} 
\usepackage{times}
\usepackage{latexsym}
\usepackage{amsmath}
\usepackage{color}
\usepackage{enumerate}
\usepackage{enumitem}
\usepackage{graphicx}
\usepackage{subfigure} 
\usepackage{float}
\usepackage{wrapfig}

\newcommand{\specialcell}[2][l]{%
  \begin{tabular}[#1]{@{}l@{}}#2\end{tabular}}

\usepackage{hyperref}

\usepackage[accepted]{icml2016}

\begin{document}

\twocolumn[
\icmltitle{Semantic Video Trailers}

\icmlauthor{Harrie Oosterhuis\footnotemark}{harrie.oosterhuis@student.uva.nl}
\icmladdress{University of Amsterdam, Amsterdam, The Netherlands}
\icmlauthor{Sujith Ravi}{sravi@google.com}
\icmladdress{Google, Mountain View, CA, USA}
\icmlauthor{Michael Bendersky}{bemike@google.com}
\icmladdress{Google, Mountain View, CA, USA}

\vskip 0.3in
]

\footnotetext{Work done at Google.}

\begin{abstract}
Query-based video summarization is the task of creating a brief visual trailer, which captures the parts of the video (or a collection of videos) that are most relevant to the user-issued query. In this paper, we propose an unsupervised label propagation approach for this task. Our approach effectively captures the multimodal semantics of queries and videos using state-of-the-art deep neural networks and creates a summary that is both semantically coherent and visually attractive. We describe the theoretical framework of our graph-based approach and empirically evaluate its effectiveness in creating relevant and attractive trailers. Finally, we showcase example video trailers generated by our system.
\end{abstract}



\section{Introduction}
\label{sec:intro}
\begin{figure*}[!bht]
\centering
\scalebox{0.8}{
\includegraphics[width=\textwidth]{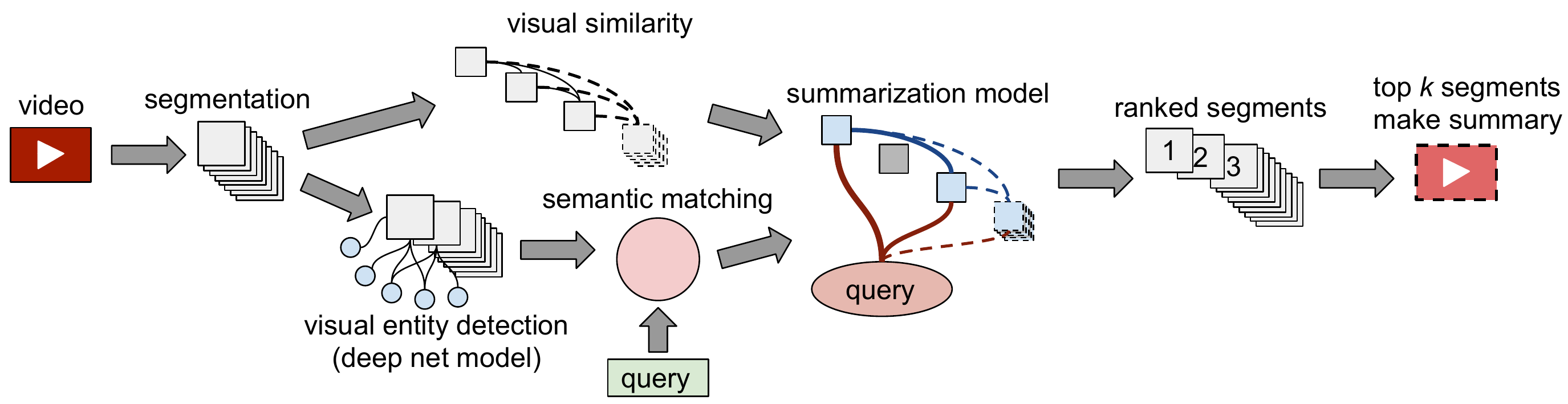}}
\caption{Graphic overview of the summarization pipeline.}
\label{fig:systemoverview}
\end{figure*}

In recent years, the availability of video content online has been growing rapidly. YouTube alone has over a billion users, and every day people watch hundreds of millions of hours on YouTube \cite{youtube2016}. With the rapid growth of available content and the rising popularity of online video platforms, accessibility and discoverability become increasingly important. Specifically, in the video search scenario, it is crucial that the platforms enable effective discovery of relevant video content.

Previous research, indeed, has dedicated a great deal of attention to video retrieval \cite{2015trecvidover}, a task that is much harder than document retrieval due to the semantic mismatch between the keyword queries and  the video frames. Therefore, video classification has been a prominent research topic \cite{karpathy2014large,brezeale2008automatic}, as well as detecting semantic concepts within video material \cite{jiang2007towards}. Both video categories and semantic concepts can be used for relevance matching between the query and parts of the video \cite{snoek2008concept}.

In this paper, we extend this existing research, and propose a system for query-based video summarization. Our system creates a brief, visually attractive trailer, which captures the parts of the video (or a collection of videos) that are most relevant to the user-issued query. For instance, for a query \emph{Istanbul}, and a video describing a trip to Istanbul, our system will construct an informative trailer, highlighting points of interest (\emph{Hagia Sophia}, \emph{Blue Mosque}, \emph{Grand Bazaar}), and skipping non-relevant content (shots of the tour bus, hotel room interior, etc.).

The applications for such a system are numerous, as such trailer skips the extraneous parts of a video, thus enhancing the user experience and saving time. For instance, it can better inform user decisions, and save time and money for services where users pay per view or pay for mobile data consumption. 
A trailer can also serve as an alternative to the standard thumbnail, a still image that represents a video in the query result list. It could potentially better capture the relevant contents of the full video than a single thumbnail image. 

The query-based summarization done by our system has two main objectives. First, the trailer will capture a \emph{semantic} match between the query and the video frames that goes beyond simple entity matching.
For instance, for a query \emph{racecar}, a frame containing a \emph{car driving on a racetrack} will be more relevant than a frame containing a \emph{stationary car}.
We achieve this semantic match via the use of entity embeddings \cite{levy2014dependency}.
Second, the trailer will be visually attractive. For instance, we will prefer frames containing visually prominent, clear depictions of relevant content. We will also prefer summaries that have smooth contiguous frame transitions, similar to human-edited movie trailers. 

The overall approach -- combining semantic match and visual similarities -- is outlined in Figure~\ref{fig:systemoverview}. In summary, the main contributions of this paper are:

\begin{enumerate}
\item A robust approach for semantically matching keyword queries to video frames, using entity embeddings trained on non-video corpora.
\item A scalable method for detecting prominent visual clusters within videos based on label propagation.
\item An efficient and effective graph-based approach that combines semantic and visual signals to construct trailers, which are both relevant and visually appealing. 
\item Detailed empirical evaluation of the proposed method with comparison to several baseline systems.
\end{enumerate}

\section{Related Work}

Previous work on video summarization has taken many different approaches to the problem and interpretations of the task. The task of summarizing a video can be interpreted as creating a textual description, a story board, a graphical representation or a \emph{video skim} that captures the content of a video appropriately \cite{money2008video}. 
In this study we address the task of constructing a \emph{video skim}, which is done by taking the video and skipping all unimportant parts. Thus all content in the resulting skim comes from the video and is played in the same chronological order. The main difference from this prior work is that our summaries are query-based.

Approaches to computing the prominence of a video fragment are widely varied. Some use only visual features, e.g. the model only adds a fragment if it is visually distinct from already added fragments \cite{zhao2014quasi,almeida2013online}. Others cluster all the frames in the video based on their visual similarity~\cite{carvajal2014summarisation}, and subsequently compose a summary by including a single fragment from each cluster. All of these approaches attempt to capture a video by covering all of its visually distinct parts.

Conversely,~\cite{gong2014diverse} propose a supervised system that learns from human created summaries. Furthermore, by using a collection of videos belonging to a very narrow category one could train a model to recognize the fragments that are the most characteristic of their category  \cite{potapov2014category}. Moreover, if no such videos are available, the model can be trained on web images of the same category \cite{khosla2013large}.
Our method contrasts with these approaches, as we incorporate a semantic interpretation of the video segments, as well as use the visual information of the fragments. In addition, our approach scales much better, as it is not restricted to a specific video category.

Existing work has also looked into using higher level concepts to construct summaries. For instance, recognizing events summaries can better address user issued event queries \cite{wang2012event}. In the same vein, detected events can be used to infer causality and construct a story-based summary \cite{lu2013story}. 

More similar to our method is previous work which recognizes ontology concepts in sports videos. A rule based method is then used to detect and include the meaningful events within the video in the summary \cite{ouyang2013ontology}. Comparable to these methods, our system computes a semantic interpretation of the video content, however we use entity embeddings, which avoids the limitation of rigid event ontologies.

Although not used for summarization, semantic embeddings have been trained for video frames. These can embody a temporal aspect as the embedding of a frame can also based on the preceding and following frames \cite{ramanathan2015learning}. Similar embeddings have been used for thumbnail detection where embeddings can be used to find the frame that is the most characteristic of the video's content \cite{liu2015multi}.
The novelty of our approach is that it uses embeddings to find the most relevant segments with respect to a keyword query and uses them for video summarization. Additionally, it is expected to create visually appealing summaries, by including visual features.

Lastly, text-based summarization methods for documents and other textual content have been long studied in the natural language processing literature. However, all these methods have primarily focused on summarizing text documents or user generated written content~\cite{submodularDispersion:2013,wangEtAl:2014}. Graph-based methods have also been used in the past for summarization~\cite{Ganesan:2010}, but in a very different context. For a detailed survey on existing text summariation techniques, see~\cite{NenkovaM12}.

\section{Method}
\label{sec:method}
In this section we propose two models for semantic query-based video summarization, the first only uses semantic information of the video whereas the second incorporates both semantic and visual information.

\begin{figure*}[tb]
\centering
\includegraphics[scale=0.22]{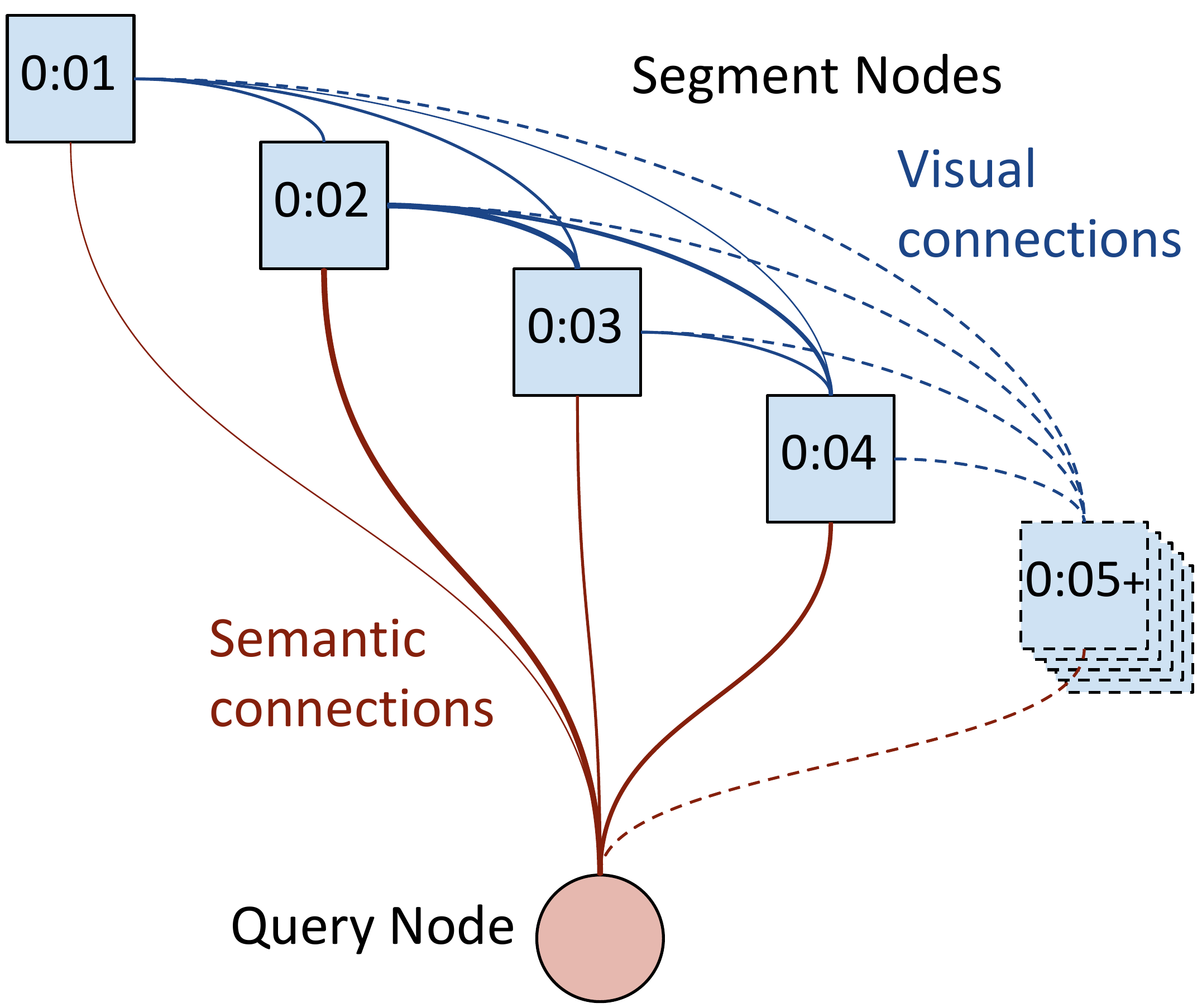} \hspace{15mm} \includegraphics[scale=0.22]{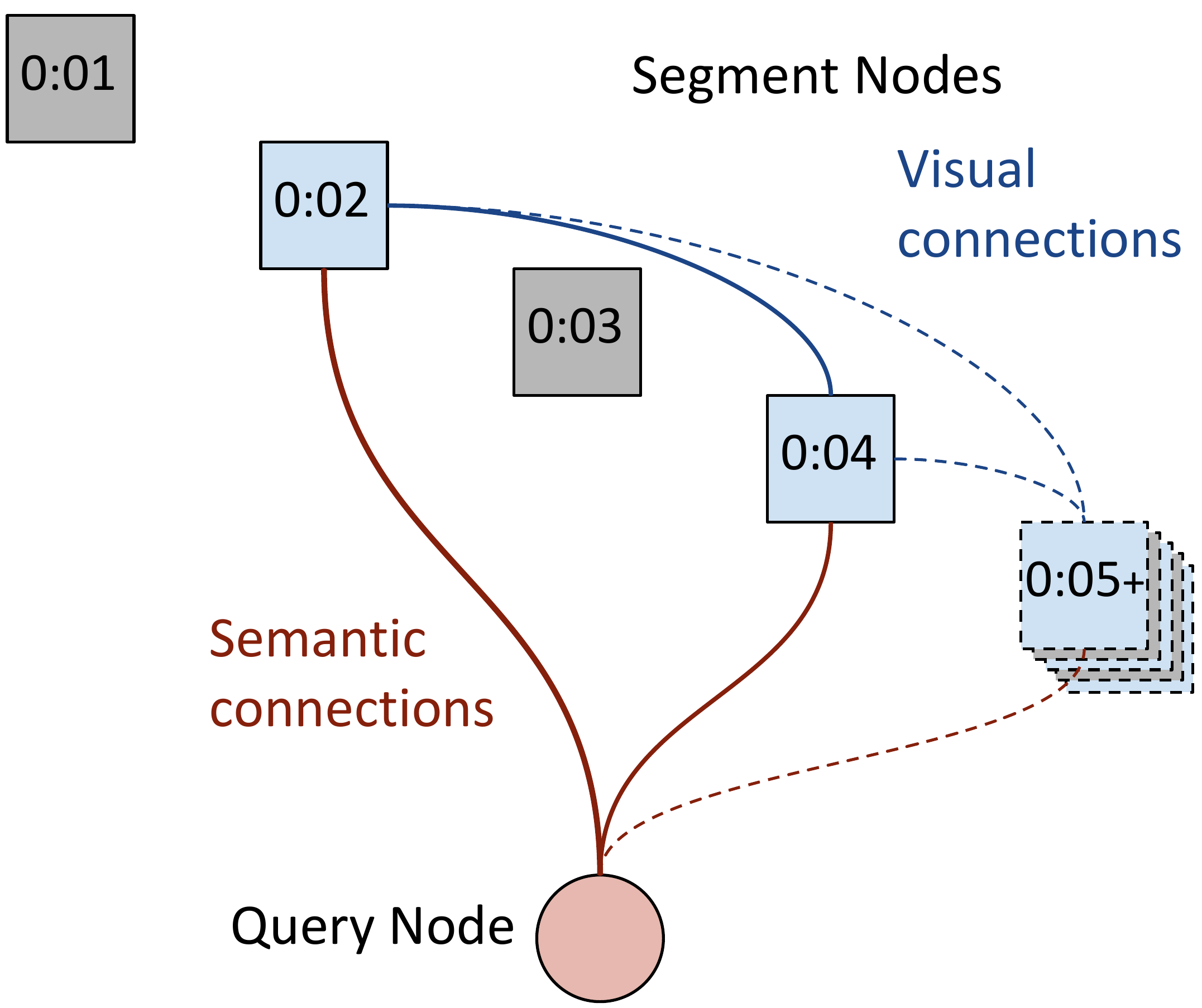}
\caption{\small  Query-video graph used for summarization before (Left) and after (Right) discarding discarding all segment nodes except for the hundred most strongly semantically connected to the query node. Query $q$ and segments $F$ from the video are represented by nodes, edges are based on visual similarity between $(F_i, F_j)$ and semantic similarity between $(q, F_i)$. For coherency all segments besides the first four have been collapsed.}
\label{fig:fullgraph}
\end{figure*}

Both models take as input a query $q$ and a video $V$; the query has been issued by a user and the video is judged to be relevant by a video retrieval algorithm. Each input video is first divided into one second segments, these are eventually used to compose the trailer summary. Working with these segments makes the final summary more comprehensible, as a second is enough time for the viewer to perceive an included clip. Furthermore, it makes the systems more scalable, as computationally expensive operations only have to be run every second instead of once for every frame in the full video. Both systems rank all the segments of a single video based on the segment content and the user query. The summary is then generated by taking the top $k=20$ ranked segments and stitching them together in order of chronological appearance in the full video. By keeping the ordering of the original video the resulting 
trailer
is expected to be more coherent, additionally the generated summary is the equivalent of a video skim. 

\subsection{Query Representation}
All our models are based on the intuition that segments capturing the same semantic content as the query should be included. Thus, the model estimates how similar the content in the query and the segment are, and ranks them accordingly. The first step in similarity estimation is to process the query $q$ and map it to a universal representation of entities $e_q \in E_q$ (and their corresponding confidence scores $w_{e_q}$), extracted from a knowledge base such as Wikipedia.

\subsection{Direct Matching}
\label{sec:method:direct}
Given the entities $E_q$ in the query, a straightforward approach is to use an image-processing model to recognize the given entities in the frame image, e.g. a deep learning architecture for concept detection in images \cite{inception,he2015spatial}. Then, the query-segment matching is simply a confidence of the concept detection model in detecting the query entities in the segment. However, this direct matching approach has several major drawbacks.

First, the number of concepts that a state of the art detection model can recognize is limited to 22,000 by the largest publicly available corpus \cite{russakovsky2015imagenet}, an extremely small subset of the entities a query can express. Moreover, processing the dataset of query-video pairs gathered for our experiments in section \ref{sec:dataset} which contains over 34,000 pairs revealed that 57\% had no entity overlap. 

Second, many summaries should contain segments that do not directly display the entities in the query but are relevant nonetheless. For instance a good summary for the entity \emph{turkey} could contain a segment of \emph{turkey stuffing} being prepared, despite that visually no \emph{turkey} is actually present. However, direct detection models are not robust enough to recognize such related concepts.

Therefore, since direct matching models cannot be applied to majority of the summarization cases, instead we focus our attention on more advanced approaches in the rest of the paper. We present two such methods next.

\subsection{Semantic Matching}
\label{sec:method:semantic}
As in the previous method, we first apply the Inception model~\cite{inception} -- state-of-the-art deep neural network architecture, that is trained to detect a large number of concepts in images -- on each frame $F_i$ in the segment. The model outputs a set of entity concepts $E_{F_i}$ with confidence scores $w_{e_f}$ for how certain the system is that each concept $e_f \in E_{F_i}$ is present in the segment $F_i$. 

However, instead of directly matching concepts between the sparse entity mappings $E_{F_i}$ and $E_q$, we compute a dense semantic embedding representation for both the query $q$ and a given video frame $F_i$ using their entity mappings. In other words, we replace each concept $e$ with its pre-computed semantic embeddings vector $\mathcal{S}_e$ . Then, a semantic representation of the segment $F_i$ is given by
$$\mathcal{S}_{F_i} = \frac{1}{|E_{F_i}|}\sum_{e_f \in E_{F_i}} w_{e_f}\mathcal{S}_{e_f}$$
Similarly, we represent the query $q$, by weighted average of embeddings for its entities to create a semantic representation $\mathcal{S}_q$.

Semantic embeddings at the entity level are computed using the recent approach from Mikolov et al.~\shortcite{mikolov13b}, and trained on a large corpus of text documents from Wikipedia. The embedding model can be learned in an unsupervised manner, thus the amount of training data can be acquired at magnitudes greater than labeled data available for training visual recognition systems. This allows the embedding model to be applicable for a substantially larger number of entities. Recent work reports 175,000 embeddings can be trained from only using the English Wikipedia \cite{levy2014dependency}.

Finally, the similarity between the query $q$ and segment $F_i$ can be estimated using the cosine similarity of their associated embeddings $\mathcal{S}_q, \mathcal{S}_{F_i}$ as follows:
\begin{align*}
\sum_{e_q \in E_{q}}\sum_{e_f \in E_{F_i}} w_{e_q}  w_{e_f} {cosine(\mathcal{S}_{e_q},\mathcal{S}_{e_f})} \\ \nonumber
= cosine(\mathcal{S}_q, \mathcal{S}_{F_i}) && \\
\end{align*}

The ranking of segments $F_i$ is based on the estimated semantic similarity to $q$, where the most similar segment is added first to the summary.

\subsection{Graph-Based Matching}
\label{sec:method:graph}

The semantic matching approach provides a robust method of estimating the relevance of segments, however it only considers semantic similarity and treats all the segments independently. Next, we introduce a second graph-based approach that models the intuition that content {\it visually} prominent in a video must be relevant to the topic it covers. In other words, besides the semantic similarity between the query and segments, the prominence of the content in a segment should also be used to estimate its relevance. We estimate prominence using visual information, thus if large parts of the video look visually similar we will assume they cover relevant content.

To effectively combine the semantic and visual signals in our system, we use Expander, an efficient graph-based learning framework based on label propagation~\cite{expander}. The framework is typically used for semi-supervised learning scenarios over graph structures~\cite{Bengio+al-ssl-2006,expander,emailcateg2016}. Usually, the weight of the edge between two nodes indicate their similarity, and true labels are known for only a subset of the nodes. The approach relies on the assumption that nodes that are very similar are also very likely to have the same labels. Accordingly the model iterates over the graph several times, at each iteration all nodes acquire the labels of the nodes they are connected to. Each node keeps a confidence score for every label based on how strongly it is connected to the nodes it acquired it from and their corresponding confident scores. In this manner, the labels are propagated through the graph at each iteration until a stable distribution of labels is reached. 
The typical use of this method is considered semi-supervised, as only a fraction of the true labels need to be known and the remaining are not learned from training data but directly inferred from the graph structure.

Our model uses a graph for each query-video pair ($q, V$) to be summarized, each segment $F_i$ extracted from the video $V$ is represented by a node in the graph, finally there is a node representing the query $q$. The values of the edges between the query node and the segment nodes are computed using the semantic matching approach, thus these edges represent their semantic similarity $cosine(\mathcal{S}_{q}, \mathcal{S}_{F_i})$. The edges between the segments on the other hand are computed by their visual similarity, this is done sampling a frame from each segment and calculating their resemblance $cosine({\mathcal{V}_{F_i}, \mathcal{V}_{F_j}})$, where $\mathcal{V}_{F_i}$ corresponds to a visual embedding corresponding to the frame $F_i$ which is computed using a hidden layer representation of the frame image within the deep learning network described earlier. A diagram of the resulting graph is displayed in Figure~\ref{fig:fullgraph}.

We learn a label assignment $\hat{L}$ on this graph that minimizes the following convex objective function:
\vspace{-0.05in}
\begin{small}
\begin{align}
\mathcal{C}(\bf \hat{L}) =&& \sum_{F_i \in V}  w_{qF_i}|| \hat{L}_q - \hat{L}_{F_i} ||_2^2  \nonumber \\  
&+& \sum_{F_i, F_j \in V} w_{ij} || \hat{L}_{F_i} -  \hat{L}_{F_j} ||_2^2 \nonumber \\ 
&+& \sum_{F_i \in V} || L_{F_i} - \hat{L}_{F_i} ||_2^2
\label{expander:obj}
\end{align}
\end{small}
where $w_{qF_i}, w_{ij}$ represent the {\it semantic} and {\it visual} similarity scores as defined above; $\hat{L}$ is the learned label distribution for query and segment nodes in the graph; and $L_{F_i}$ is the seed label (i.e., identity) on the video segment nodes.

The segment nodes are each assigned a unique ``seed'' label (i.e., their identity). We optimize the above objective function using the iterative streaming algorithm described in \cite{expander}, then after running label propagation the confidence scores of the labels acquired by the query node $\hat{L}_q$ are considered. The segments are ranked corresponding to how strongly their corresponding labels were propagated to the query node. In other words, the output label scores on the query node $\hat{L}_q$ indicate how well the segments are connected to the query in the graph. A segment can be strongly connected because it is semantically similar to the query or it is visually similar to other segments that are strongly connected. Note that contrary to the typical usage of label propagation, our approach is in fact unsupervised as the initial labels can automatically be assigned. The streaming Expander algorithm permits efficient scaling to thousands or millions of frames for long videos while maintaining constant space complexity.

Presumably we could ignore the semantic edges in the graph completely and propagate only the frame-ids over the visual edges. This is equivalent to performing visual clustering, we do not consider this model here because it ignores the query and therefore is unsuited for this task. Similarly the  edges could be weighted so that the model values the either semantic or visual signals more. We can also easily incorporate {\em diversity} among ranked results, as in traditional summarization approaches, by simply converting the visual similarity signal into a distance metric.\footnote{Different graph configurations were tried but are not included in to maintain brevity.} Furthermore the generic setup of the method allows it to be easily extended with novel signals in the future.

Though the intuition behind the previous graph construction is reasonable, preliminary results revealed some practical problems with this model. Namely many videos contain visuals that often recur in the video but are not relevant for a summary. For instance, news shows or documentaries can feature a presenter who talks periodically throughout the video. These segments will be very similar visually despite being the least interesting parts to include in a summary. Moreover this problem can be extremely prevalent in online video content, since they often feature an almost static outro where users are asked to leave favorable feedback and watch more videos. Because these outros usually consist of text on a near static background, they form very strong clusters in the graph which boost these segments into the summary.

To counter these issues, we change the model to instead only consider the hundred highest semantically similar segments, thereby yielding a {\it graph-based reranking} model. The nodes representing the other segments and their edges are completely disregarded, as can be seen in the Figure~\ref{fig:fullgraph}. The intuition behind this reranking model is that content prominent among the relevant parts of a video are expected to be good additions to a summary and the irrelevant frames are automatically discarded.

\section{Experiments}

\label{sec:experiments}
In this section, we detail our experiments designed to evaluate the performance of our models.  Section~\ref{sec:baselines} introduces two baselines for comparison, subsequently we discuss the data used for evaluation and our experimental setup in Section~\ref{sec:dataset} and Section~\ref{sec:setup} respectively.

\subsection{Baselines}
\label{sec:baselines}
To  properly investigate the performance of the models introduced in Section~\ref{sec:method} we introduce the {\bf uniform} baseline model for comparison. Similar to the models, the uniform baseline also uses one second segments, however instead of judging their relevance the method selects segments according to a uniform distribution.
As a result each segment is equally likely to appear in the generated summary.
Because the uniform sampling covers all parts of the video equally, the summary is expected to capture all parts of the video. Since the video is selected using a state-of-the-art retrieval method, its content is expected to be very relevant to the topic. Thus the resulting summary is expected to be just as relevant to the query. 
However since it does not take into account the content of the video nor the query, 
it is expected to fail on videos that spend disproportionate time on some topics or contain cover material unrelated to the query. Both of these are unlikely if a strong retrieval model was used or if it was a short video.

Additionally, we introduce a second baseline: the first twenty seconds model ({\bf first-20}). This baseline creates a summary of a video by taking its first twenty seconds. This simple model is based on two intuitions. 
Firstly the generated summaries keep the coherency of the original video because each summary is an unaltered clip where no \emph{film cuts} were introduced. Secondly, many videos start with an introduction of their topic usually to gain the viewers attention. Accordingly, this baseline attempts to select a single clip that gives an overview of the video.

\subsection{Dataset}
\label{sec:dataset}

Since our proposed system uses a query and a matching video, we make use of YouTube to collect these query-video pairs. Because YouTube receives millions of user queries per day and has a large variety of content, we consider it a good fit to test the effectiveness of our system.
We sampled 1800 of the most commonly issued queries, for each query twenty matching videos were sampled uniformly from the top hundred search results. Subsequently the summarization system was then applied to the resulting 34,725 videos, note that some videos are matched to multiple queries.

Sampling of videos was limited to those with a running length greater than ten minutes. This makes sure that summarization is not a trivial task. 
In addition, video-query pairs which had an overlap in extracted entities were discarded as well. We chose to discard these videos to test the robustness of our system, since this limitation makes the direct match approach (described in Section~\ref{sec:method:direct}) impossible. As a result the data only contains instances where the semantic similarity between segments and the query cannot be computed directly. As described in Section~\ref{sec:method:semantic} our system can handle these entity mis-matches by using semantic embeddings. We believe this focus on the mis-matching cases is warranted, as we consider wide applicability as more important than good performance on a particular video subset. 

Lastly since the system was evaluated using crowdsourcing we were unable to use the entire set of summarized query-video pairs. Instead a subset of 127 query-video pairs was used for the crowdsourced evaluation.

\subsection{Experimental Setup}
\label{sec:setup}

The quality of a summary is difficult be judged objectively. Consequently we used the Amazon Turk platform to perform a crowdsourced experiment, with three raters per task. Our comparison of models and baselines is based on the crowdsourced assessments of generated summaries.

However the task of judging a single summary proved to be very hard for most people, instead we found asking for preferences between summaries is a more comprehensible task. Accordingly the task consisted of a single question: ``\emph{Someone is looking for a video about }[query], \emph{which of the following two 20 second videos is best to show?}" followed by two side-by-side summary trailers: one generated by a model, and another by a baseline, their order randomized.

A judgement was collected for the combination of each query-video pair, model and baseline, giving us a total of 508 judgments.
However we noticed that some users disregarded the task to quickly optimize on the money incentive. For this reason we disregarded any judgement made within less than 30 seconds, bringing the number of judgements down to 449. Significance testing of the preferences between the systems was done by applying a two sided Wilcoxon sign test.

\begin{table}[bth]
\small
\begin{tabular*}{\columnwidth}{ l r r }
\toprule
\bf Model & \bf Pref. over first-20 & \bf Pref. over uniform \\
\midrule
\multicolumn{3}{c}{\it All videos}\\
\midrule
semantic & 74\% & 50\% \\
graph-based & 73\% & 56\% \\
\midrule
\multicolumn{3}{c}{\it Gaming and animation categories}\\
\midrule
semantic & 76\% & 48\% \\
graph-based & 74\% & 52\% \\
\midrule
\multicolumn{3}{c}{\it Non gaming and animation categories}\\
\midrule
semantic & 73\% & 51\% \\
graph-based & 72\% & 58\% \\
\midrule
\multicolumn{3}{c}{\it Videos under 20 minutes}\\
\midrule
semantic & 73\% & 43\% \\
graph-based & 84\% & 56\% \\
\midrule
\multicolumn{3}{c}{\it Videos of 20 minutes and over}\\
\midrule
semantic & 75\% & 56\% \\
graph-based & 64\% & 55\% \\
\bottomrule
\end{tabular*}
\vspace{0pt}
\caption{\small Results of the experiment described in Section~\ref{sec:experiments}. Percentages show preference of the summaries of one system over that of the baseline.}
\label{tab:preferences}
\end{table}

\section{Results}
\label{sec:results}

In this section, we present the results of our experiment described in Section~\ref{sec:experiments}, provide several example summaries and evaluate our proposed summarization method.

\subsection{Experimental results}
\label{sec:experimentalresults}

The results of our crowdsourcing experiment are displayed in Table~\ref{tab:preferences}. A clear preference of both models over the first twenty seconds baseline is visible. Since they are statistically significant ($p~<~0.01$) we conclude that both our models create better summaries than this baseline. When compared to the uniform baseline though, the graph-based approach yields more favorable summaries compared to the semantic-only model. However, overall preference \% for the two models compared to the uniform baseline are not as high. There could be several reasons for this, e.g., the task is not easy for people who are not familiar with video summarization. 

Furthermore, the videos may not be appropriate for summarization; to further investigate this judgements were split based on video-category and length. Table~\ref{tab:preferences} shows the preferences for videos in the Gaming and Animation category (29\% of videos) and all others. These categories were chosen as they are prevalent on YouTube and are expected to be less suited for summarization.
The results show us that both models perform better for Non Gaming and Animation categories when compared to the uniform baseline. Additionally, results split by video length are also displayed in Table~\ref{tab:preferences}, we chose to split on 20 minutes as close to half (44\%) are under 20 minutes. Here we see that the semantic model performs substantially better on videos over 20 minutes with a 13\% difference compared to the uniform model, though graph-based performs almost the same with a 1\% difference. These results suggest that certain types of videos are more suited for auto-generating summary trailers.

 In addition to the previous experiment, we performed a more detailed study on a smaller video dataset to better understand the differences between models. This experiment was also crowdsourced and showed judges a single summary together with a multi-choice questionnaire; videos were sampled and judgements were gathered for their summaries created by the uniform baseline and the graph-based model. In total 60 judgements were collected, the questions and results are displayed in Table~\ref{tab:questionnaire}, answers ranged from 1 (most negative) to 5 (most positive). The questionnaire shows us a clear signal that the graph-based method creates summary trailers that are visually more attractive than the uniform baseline.
 \vspace{-1em}
 \begin{table}[bth]
\small
\begin{tabular*}{\columnwidth}{ l c c c }
\toprule
\bf Question & \bf uniform & \bf \specialcell{graph\\-based} & \bf $\Delta$ \\
\midrule
\specialcell{Rate the visual quality\\of the summary,\\how good does it look?} & 3.54 & 3.94 & +11.16\% \\
\midrule
\specialcell{For query X, how well\\does the summary\\capture all relevant parts\\of the video?} & 4.38 & 4.47 & +1.87\% \\
\midrule
\specialcell{For query X, how relevant\\is the summary?} & 4.15 & 4.27 & +2.72\% \\
\bottomrule
\end{tabular*}
\vspace{-2.4pt}
\caption{\small Average results of questionnaire, scores range from 1 (most negative) to 5 (most positive).}
\label{tab:questionnaire}
\end{table}

\begin{figure}[ht!]
\centering
\includegraphics[width=\columnwidth]{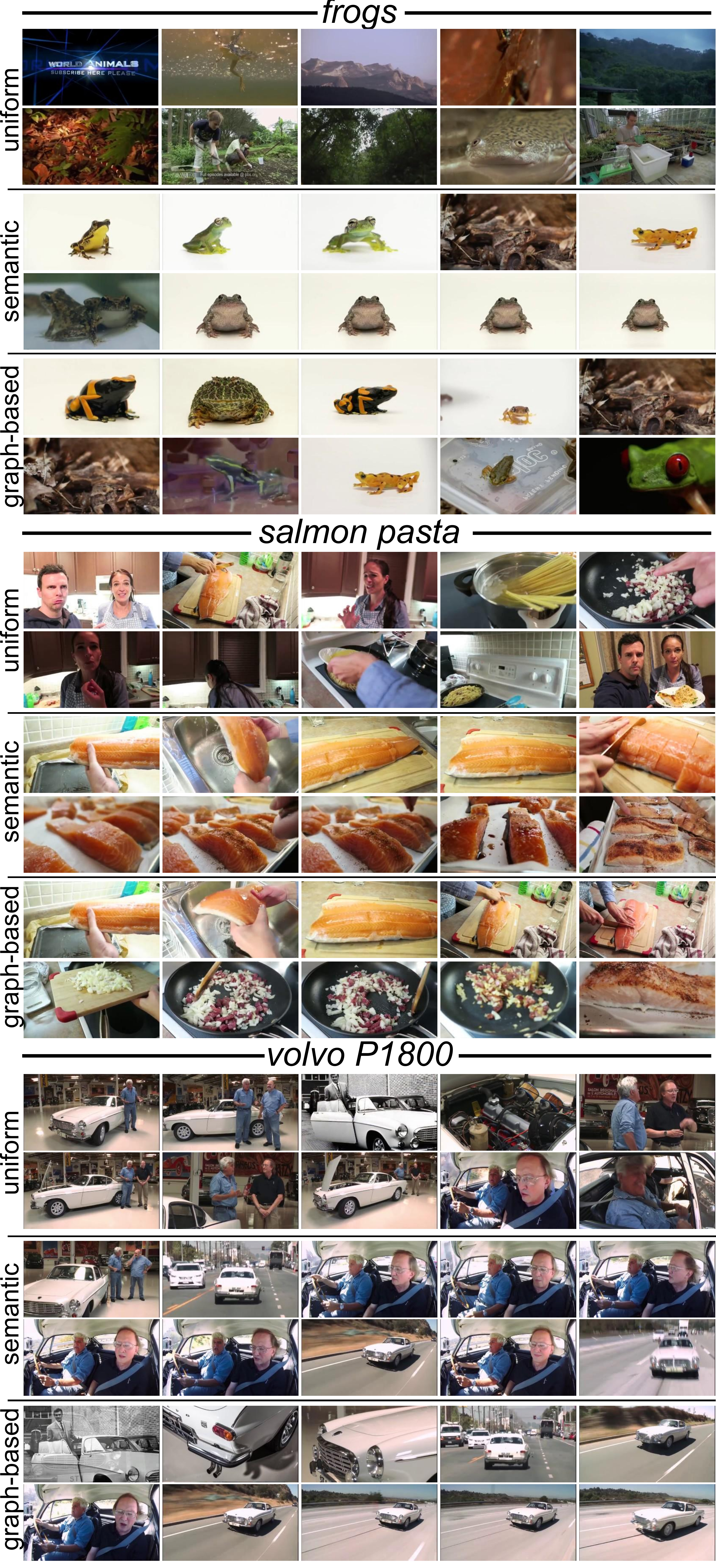}
  \vspace{-1.9em}
\caption{\small Summaries created by the uniform, semantic and graph-based models for the queries: \emph{frogs}, \emph{salmon pasta} and \emph{volvo P1800}. Visualized by sampling a frame every 2 seconds.}
  \vspace{-2.4em}
\label{fig:videoskim}
\end{figure}

\vspace{-1.8em}
\subsection{Example summaries}
\label{sec:examplesummaries}

To further investigate the effects of using different models we display example summaries in Figure~\ref{fig:videoskim} which are the result of applying different models to the same three query-video pairs.
For this illustration the uniform baseline, semantic model and graph-based model were applied, the first twenty seconds baseline was dismissed as 
it performs significantly worse according to the results in Section~\ref{sec:experimentalresults}.
Three videos were sampled from different categories to illustrate robustness and diversity, the selected query-video pairs are: \emph{frogs}, an animal documentary; \emph{salmon pasta}, an amateur cooking video; \emph{volvo P1800}, an informational video regarding a famous car model\footnote{Videos are available under the Creative Commons licence at: youtu.be/w-AItfioqlw, youtu.be/tR9ZtaGtCAM and youtu.be/FwCjOakOMKE}.

The uniform summaries cover the videos passably, however the summaries contain many shots unrelated to the query. Most notably all uniform summaries contain shots of people who are presenting the video but are not relevant to the query.
In contrast, the semantic summaries only contain shots related to the query. For the first video we see that the semantic model has only included shots containing frogs, for the \emph{salmon pasta} video only shots of fish are included, and for the \emph{volvo P1800} video the summary consists of only shots that clearly display cars. Therefore we conclude that the semantic model can recognize semantic similarity robustly, as it found relevant shots effectively despite the fact that no direct annotations of the query were available in the video.

Lastly we have the graph-based summaries, as expected they are very similar to those of the semantic model. The differences are important though: the \emph{frogs} summary displays more shots of more different frogs, which adds diversity to the video. The model picked up on shots where the frogs are less directly recognizable (for instance due to camouflage or displaying the head) due to their visual similarity to semantically relevant shots. In the \emph{salmon pasta} summary shots of the vegetable sauce are included, the model inferred their relevance due to their prominence in the video. The semantic model did not include these as \emph{salmon pasta} is defined by its fish, however with respect to the cooking video this seems to be a good inclusion. Finally, the \emph{volvo P1800} summary displays more shots showing the outside of the car. The model picked up on interesting shots by their prominence and the result is a more visually appealing summary.

These examples show a clear difference between the uniform baseline and our models. This contrasts with some of the results in Section 5.1, where the preference differences between our models and the uniform model were not as pronounced. This suggests that the query-based video summarization task is a difficult one, and visual summary evaluation is an interesting direction for future work.

\section{Conclusion}

We presented a system for query-based video summarization that effectively combines semantic interpretations and visual signals of the video to construct summary trailers.

Despite the difficulties of evaluating for this complex task, we show that the new approach outperforms other baselines in terms of summarization quality as judged by human raters. We also show several examples which demonstrate that the approach of combining embeddings with frame annotations allows for robust semantic detection of relevant segments. 

Moreover, our proposed graph-based model is able to recognize parts of the video that are both relevant to the query and visually prominent in the video. Future research could expand this approach by applying the graph-based model over several related videos to find latent topics using their visual similarity or to create multiple summary views per video each focused on a different topic. Finally, the usage of query-based summaries as dynamic thumbnails seems a promising direction for research.

\balance
\bibliography{arxiv2016-trailer}
\bibliographystyle{icml2016}

\end{document}